\documentclass[letterpaper, 10 pt, journal, twoside]{ieeetran}
\usepackage{amsmath,amssymb,amsfonts}
\usepackage{algorithmic}
\usepackage{algorithm}
\usepackage{graphicx}
\usepackage{caption}
\usepackage{array}
\usepackage{stfloats}
\usepackage{url}
\usepackage{cite}
\usepackage[caption=false,font=normalsize,labelfont=sf,textfont=sf]{subfig}
\usepackage{textcomp}
\usepackage[dvipsnames]{xcolor}
\usepackage{booktabs}
\usepackage[final]{changes}
\definechangesauthor[name={Yuta Shimane}, color=red]{*}
\definechangesauthor[name={Yuta Shimane R2}, color=red]{**}

\usepackage{makecell}

\newcommand{\bm}[1]{\mbox{\boldmath $ #1 $}}

\newif\ifshowRtwo
\showRtwotrue  

\usepackage{ifthen}

\usepackage[absolute,overlay]{textpos}

\newif\ifshowRtwo
\showRtwotrue    

\ifshowRtwo

  \newcommand{\addedRone}[1]{#1}
  \newcommand{\deletedRone}[1]{}
  \newcommand{\replacedRone}[2]{#1}
  
  \newcommand{\addedRtwo}[1]{\added[id=**]{#1}}
  \newcommand{\deletedRtwo}[1]{\deleted[id=**]{#1}}
  \newcommand{\replacedRtwo}[2]{\replaced[id=**]{#1}{#2}}
\else

  \newcommand{\addedRone}[1]{\added[id=*]{#1}}
  \newcommand{\deletedRone}[1]{\deleted[id=*]{#1}}
  \newcommand{\replacedRone}[2]{\replaced[id=*]{#1}{#2}}
  
  \newcommand{\addedRtwo}[1]{#1}
  \newcommand{\deletedRtwo}[1]{}
  \newcommand{\replacedRtwo}[2]{#2}
\fi

\IEEEoverridecommandlockouts                              

\begin{document}

\title{
Simulation of Adaptive Running with Flexible Sports Prosthesis \\ using Reinforcement Learning of Hybrid-link System
}

\author{Yuta Shimane$^{1}$ and Ko Yamamoto$^{2}$
\thanks{Manuscript received: November 23, 2025; Revised: April 7, 2026; Accepted: April 22, 2026.}
\thanks{This paper was recommended for publication by
Editor Pietro Valdastri upon evaluation of the Associate Editor and Reviewers
comments. This work was supported by the JSPSKAKENHI under Grant 21H01282.}
\thanks{$^{1}$
Yuta Shimane is with Department of Biological Sciences, The University of Tokyo, Bunkyo-ku, Tokyo, 113-0033, Japan {\tt\footnotesize yuta-shimane@g.ecc.u-tokyo.ac.jp}
}%
\thanks{$^{2}$
Ko Yamamoto is with Institute of Systems and Information Engineering, University of Tsukuba, Ibaraki, 305-8573, Japan {\tt\footnotesize yamamoto@iit.tsukuba.ac.jp}
}%
\thanks{Digital Object Identifier (DOI): https://doi.org/10.1109/LRA.2026.3693933}
}

\markboth{IEEE Robotics and Automation Letters. Preprint Version. Accepted May, 2026}
{Shimane \MakeLowercase{\textit{et al.}}: Simulation of Adaptive Running with Flexible Sports Prosthesis using RL of Hybrid-link System} 

\maketitle

\begin{abstract}
This study proposes a reinforcement learning–based framework for adaptive running motion simulation in a unilateral transtibial amputee using a hybrid-link system that incorporates the flexibility of a leaf-spring-type sports prosthesis.
The design and selection of sports prostheses typically rely on trial and error.
A comprehensive whole-body dynamics analysis that accounts for interactions between human motion and prosthetic deformation can provide valuable insights for user-specific design and selection.
The proposed hybrid-link system enables such analysis by integrating a Piece-wise Constant Strain (PCS) model to represent prosthetic flexibility.
Based on this system, the simulation methodology generates whole-body dynamic motions of a unilateral transtibial amputee using a reinforcement learning approach. This framework integrates imitation learning based on motion capture data with accurate computation of prosthetic dynamics.
Running motions are simulated under multiple virtual prosthetic stiffness conditions, and the corresponding metabolic cost of transport (COT) obtained from these simulations is analyzed. 
The results suggest that variations in prosthetic stiffness influence running dynamics and performance, and that COT is consistent with values reported in prior study.
Our findings demonstrate the potential of the proposed approach for simulation and analysis under virtual conditions that differ from real-world conditions.
\end{abstract}


\begin{textblock*}{18cm}(1.8cm,26.7cm)
\footnotesize © 2026 IEEE.  Personal use of this material is permitted.  Permission from IEEE must be obtained for all other uses, in any current or future media, including reprinting/republishing this material for advertising or promotional purposes, creating new collective works, for resale or redistribution to servers or lists, or reuse of any copyrighted component of this work in other works.
\end{textblock*}


\section{Introduction}
\label{sec:intro}
\IEEEPARstart{P}{rostheses} designed specifically for running \replacedRtwo{have}{has} contributed to the achievements of parasports players, and their performance has improved annually with advances in prosthetic research and development.
However, the design and selection of prostheses often rely on trial-and-error processes, which suggests that individuals may not always use the most suitable prosthesis for their specific requirements.
This limitation arises from the wide variety of available prosthetic limbs, with factors, including material properties, geometric configurations, as well as mounting height and angle, influencing sports performance \addedRtwo{\cite{beck2017, murai2018, guzelbulut2021}}.
Accordingly, enabling whole-body dynamic simulations that explicitly model the characteristics associated with prosthetic deformation would provide valuable insights for biomechanical analysis and support the selection of an appropriate sports prosthesis.

\begin{figure}[t]
  \centering
    \includegraphics[width=1.0\hsize]{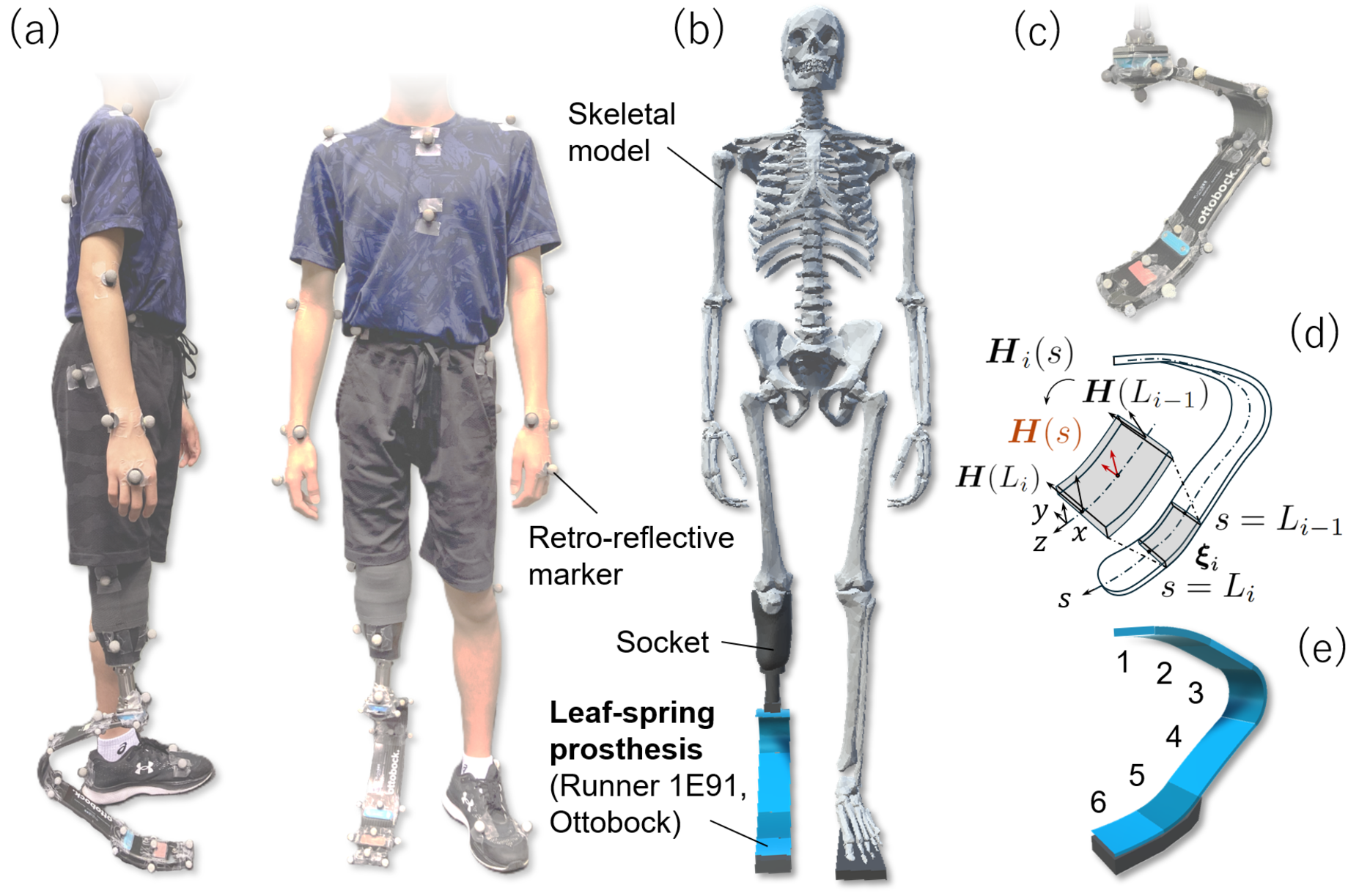}
  \caption{Overview of the prosthesis integrated skeletal model and motion measurement: (a) Measurement \replacedRone{setup using optical motion capture.}{ of the unilateral transtibial amputee was performed using optical motion capture and force plates.} (b) Integrated prosthetic skeletal model using the hybrid-link system. \replacedRone{The model has 33 DOFs including 9 DOFs for the skeleton and 18 DOFs for the prosthesis.}{ : The model has 33 DOFs, including 9 DOFs for the skeleton and 18 DOFs for the prosthesis.} (c) Leaf-spring-type sports prosthesis.\deletedRone{, Runner 1E91, Ottobock} (d) PCS model \replacedRone{for flexible rod.}{ : The rod shape of the prosthesis is divided into a finite number of segments, and the deformation is defined as the displacement of the configuration curve.} (e) PCS model of the prosthesis \replacedRone{comprising 6 segments.}{ : The prosthesis comprises 6 segments of the PCS model. } }
  \label{fig:model}
  \vspace{-2.0em}
\end{figure}

Human musculoskeletal models \cite{nakamura2005, Delp2007} are employed to calculate whole-body motion.
Several studies have incorporated prosthetic flexibility into musculoskeletal simulations using linear springs \cite{murai2018} or rigid multi-link systems with passive joints \cite{guzelbulut2021, rigney2016}.
Although these approaches offer useful insights into the effects of prosthetic flexibility, their reliance on simplified representations or rigid-body assumptions limits the accuracy of deformation modeling and may obscure critical biomechanical factors in the analysis.
Accurate evaluation of the mechanical effects of prosthetic deformation therefore requires modeling approaches that reflect the actual material properties of the prosthesis.

The finite element method (FEM) is widely used for modeling flexible structures, and a framework for developing complex and high-performance simulators has been proposed \cite{faure2012}.
FEM has also been applied for the assessment \cite{rigney2017} and design \cite{Shepherd2023} of prostheses, such as for in vitro mechanical testing at various orientation angles.
However, such modeling approaches do not capture the dynamics and time-varying deformation behavior of the prosthesis during running. 
Furthermore, although FEM can represent complex geometries with high fidelity, the associated computational cost increases with the number of mesh elements.
Additionally, flexible deformation modeling such as the Piece-wise Constant Strain (PCS) model \cite{renda2018} has been explored in soft robotics.
The PCS model enables a computationally efficient analysis compared with FEM.
Furthermore, as the PCS model shares structural similarities with the rigid multi-link systems, it is effective for analyzing the interaction between flexible prosthetic structures and body movement.
Accordingly, a hybrid-link system that integrates the PCS model with a rigid multi-link system has been proposed \cite{ishigaki2025_, Mathew2023}.
This hybrid-link system has been applied to model prostheses in combination with skeletal models and to analyze running motions \cite{kim2026}.
In addition, prosthetic viscoelastic properties based on the PCS model have been estimated using motion capture data and ground reaction forces, and their accuracy has been validated through forward dynamics simulations
\cite{shimane2024}.
These approaches enable accurate modeling of prosthetic behavior and facilitate extension to whole-body forward dynamics simulations \addedRone{for investigating how prosthetic conditions affect running performance}.
\begin{table}[t]
\footnotesize
\centering
\setlength{\tabcolsep}{2pt}
\renewcommand{\arraystretch}{0.95}

\caption{\addedRone{Comparison of related works}}
\label{table:works}
\vspace{-0.5em}

\begin{tabular}{llllll}
\toprule

 \shortstack[c]{Related\\works}
& \shortstack[c]{Whole-body\\computation}
& \shortstack[c]{Body\\dim.}
& \shortstack[c]{Pros.\\dim.}
& \shortstack[c]{Pros.\\Modeling}
& Key feature \\

\midrule

\cite{rigney2017, Shepherd2023}
& -- & -- & 3D & FEM 
& \shortstack[l]{High fidelity; expensive} \\

\cite{rigney2016, guzelbulut2021}
& ID & 3D & 2D & Rigid-link 
& \shortstack[l]{Efficient; no deform} \\

\cite{kim2026}
& ID & 3D & 3D & PCS 
& \shortstack[l]{Deform; no simulation} \\

\cite{hase2020}
& FD+PD & 2D & 2D & Flat spring 
& \shortstack[l]{Efficient; limited control} \\

\cite{murai2018}
& FD+SLIP & 3D & 2D & Rigid-link 
& \shortstack[l]{Generative; simplified} \\

[16]-[18]
& FD+RL & 3D & -- & -- 
& Learning-based control \\

\midrule

\textbf{This study} 
& \textbf{FD+RL} & \textbf{3D}* & \textbf{3D} & \textbf{PCS} 
& \shortstack[l]{\textbf{Physically grounded;}\\ \textbf{not expensive}} \\

\bottomrule
\end{tabular}

\vspace{2pt}
\footnotesize
\textit{Note:} Pros. = prosthesis. *Pelvis link constrained to the sagittal plane.
  \vspace{-3.0em}
\end{table}

\replacedRone{Table \ref{table:works} summarizes related works on prosthetic motion analysis.
Whereas inverse dynamics (ID) was employed in \cite{guzelbulut2021,rigney2016,kim2026}, forward dynamics (FD)}
{
Forward dynamics
} simulations compute the time evolution of joint and prosthetic motion under prescribed forces and initial conditions.
Therefore, simulations conducted under prosthetic conditions that differ from those observed in real-world settings have the potential to provide important guidelines for the selection and design of sports prostheses.
This requires modeling the human motor control system that governs skeletal dynamics. 
\replacedRone{Although \cite{murai2018,hase2020} employed spring inverted pendulum model or PD control, these approaches were too simple to model the human motor control system.
Moreover, the model of the prosthesis in \cite{murai2018,hase2020} was relatively simple or limited to 2D motion.}{
Walking behavior can be modeled using neural dynamics associated with rhythmic pattern generation, as well as through optimal control–based approaches.
}
\replacedRone{Recently, reinforcement learning (RL)}{Reinforcement learning–based simulation} techniques have enabled the control of complex musculoskeletal models \cite{weng2021}.
In the context of imitation learning under behavior cloning, the control policy is learned directly from expert demonstration data. 
In addition, by designing reward functions that explicitly encourage the agent to reproduce human motion, imitation learning–based RL can achieve faster convergence toward optimal behavior \cite{peng2018, Han2022}, although these approaches have primarily been applied to rigid multi-link systems.
These learning-based frameworks enable simulations of both prosthetic and human dynamics under a wide range of design parameters, including stiffness, shape, and mounting conditions.

In this study, 
\addedRone{we construct a motion analysis framework that integrates FD simulation of a hybrid-link system with a RL–based control policy.
The hybrid-link system enables precise modeling of the flexible deformation behavior of the prosthesis while maintaining a reasonable computational cost, and its dynamical computation provides a quantitative analysis that accounts for the physical interaction between the human limbs and the prosthetic leg.
Moreover, RL-based control policy enables the reproduction of subject-specific motion using the actual prosthesis, while also allowing the generation of {\it virtual} motion under hypothetical conditions in which the same subject is assumed to use prostheses with different parameter settings.
Accordingly, the research questions addressed in this study are defined as follows:}
    \begin{enumerate}
        \item \addedRone{Is it possible to generate dynamic motion simulations by learning a control policy that reproduces the motion of an actual transtibial amputee?}
        \item \addedRone{When the characteristics of the prosthesis differ from those in the actual condition, to what extent are the simulation results consistent with results reported in the literature?}
    \end{enumerate}
\deletedRone{we present a RL-based methodology for a hybrid-link system that simulates running motion in a unilateral transtibial amputee with a flexible prosthesis.
Although the kinematics and dynamics of hybrid-link systems have been studied, their control through RL remains largely unexplored.}
\addedRtwo{In particular, for the second research question, the simulated running behavior is evaluated using the metabolic cost of transport (CoT), as investigated in \cite{beck2017}. This metric serves as a key indicator of athletic performance and plays an important role in running.}

The remainder of this paper is organized as follows.
Section \ref{sec:hybridlink} describes the modeling of unilateral transtibial amputee motion using the hybrid-link system, together with the motion capture setup.
Section \ref{sec:RL} presents the RL framework, including the definitions of the state, action, reward, and network architecture.
Section \ref{sec:imitation} describes the implementation of RL using parallel computing and fine-tuning, along with the resulting walking and running motions generated by the model.
Section \ref{sec:effect} presents simulation results obtained under different prosthetic stiffness conditions and discusses the associated energy consumption.
Finally, \addedRone{Section \ref{sec:discussion} discusses the obtained results, and} Section \ref{sec:conclusion} concludes the paper.
\section{Hybrid-link System Integrating Skeletal System and Leaf-Spring-Type Prosthesis}
\label{sec:hybridlink}

\subsection{PCS Model \cite{renda2018} for Leaf-Spring-Type Prosthesis}
\label{subsec:PCS}
Fig. \ref{fig:model} provides an overview of the subject and hybrid-link model, in which the PCS model \cite{renda2018} is used for modeling the prosthesis, as shown in Fig. \ref{fig:model} (d).
%
The coordinate $s$ is defined along the center axis of the rod.
The configuration curve of the rod can be represented by the homogeneous transformation matrix $\bm{H}(s) \in \rm{SE}(3)$.
Using the spatial derivative of $\bm{H}(s)$, the deformation vector 
$\bm{\xi}(s) \in \mathbb{R}^6$ 
is defined as follows:
\begin{align}
\label{xi}
[\bm{\xi}(s)\times] :=\bm{H}^{-1} \frac{\partial \bm{H}}{\partial s}
\in \rm{se}(3),
\end{align}
In this paper, $\bm{\xi}(s)$ is referred to as the {\it strain} vector.
Hereinafter, $s$ is omitted when its meaning is clear from the context.

We divide the rod into $N$ segments and assume that the strain is constant along the central axis coordinate $s$ within each segment.
Therefore, the $i$-th segment $(i=1,\cdots,N)$ is defined as $L_{i-1} \le s < L_{i}$, and the strain of the $i$-th segment is defined as follows:
\begin{align}
    \label{eq:cnst_strain}
 \bm{\xi}_i:= \bm{\xi}(s) \ (L_{i-1} \le s < L_{i}).
\end{align}

We define $\bm{H}_{i-1} := \bm{H}(L_{i-1})$, which is the configuration curve at the starting position of segment $i$: $s = L_{i-1}$.
Given $\bm{\xi}_i$, the configuration curve of segment $i$ is calculated as follows:
\begin{align}
    \label{eq:pcs_fk01}
    \bm{H}(s) = \bm{H}_i \exp\{ (s - L_{i-1}) [\bm{\xi}_i \times]\}
     \ (L_{i-1} \le s < L_{i}).
\end{align}
Given the strain values for all segments, the configuration curve of the rod can be calculated by recursively applying \eqref{eq:pcs_fk01}
to $i = 1, \cdots, N$.
Therefore, the generalized coordinate $\bm{q}_\mathrm{S} \in \mathbb{R}^{6N}$ of the PCS model is defined as follows:
\begin{align}
 \bm{q}_\mathrm{S} =
\begin{bmatrix}
 \bm{\xi}_1^T & \bm{\xi}_2^T & \cdots & \bm{\xi}_N^T
\end{bmatrix}^T.
\end{align}

\addedRone{We modeled the prosthesis (Runner 1E91, Ottobock\addedRone{, Germany}) as a PCS model with a total of 18 DOFs, as shown in Fig.~\ref{fig:model} (e), following the procedure presented in \cite{shimane2022}.}

\subsection{Hybrid-Link System Dynamics}
\label{subsec:Hybrid}
We model the whole-body motion of a prosthetic runner using a hybrid-link system \cite{ishigaki2025_, Mathew2023}, 
as shown in Fig. \ref{fig:model} (b).
Hereafter, the subscripts $\mathrm{R}$ and $\mathrm{S}$ represent the rigid-link system and the PCS model, respectively.
The equation of motion for a floating-base system
is expressed as follows:
\begin{align}
    &\bm{M}\ddot{\bm{q}} + \bm{b} = \bm{\tau} + \sum_{i} \bm{J}_{\mathrm{C}, i}^T \bm{f}_{\mathrm{C}, i},
\end{align}
\begin{itemize}
    \item The generalized coordinates and velocities of the hybrid-link system are defined as follows:
    \begin{align}
        \bm{q} &:= 
        \{
        \bm{H}_0,  \bm{q}_\mathrm{R}, \bm{q}_\mathrm{S}
        \}, \quad \nonumber
        \dot{\bm{q}} := 
        \begin{bmatrix}
            \bm{\eta}_0^T \  \dot{\bm{q}}_\mathrm{R}^T \  \dot{\bm{q}}_\mathrm{S}^T
        \end{bmatrix}^T,
    \end{align}
    where $\bm{H}_0 \in {\rm SE}(3)$ represents the position and orientation of the base link, and $\bm{\eta}_0$ denotes its spatial velocity. 
    \item $\bm{M}$ is the inertia matrix of the hybrid-link system.
    \item $\bm{b}$ represents the bias term that summarizes the effects of the centrifugal force, Coriolis force, and gravity.
    \item $\bm{\tau}$ is the generalized force, defined as follows: 
    \begin{align}
        \bm{\tau} = [\bm{\tau}_0^T \  \bm{\tau}_\mathrm{R}^T \  \bm{\tau}_\mathrm{S}^T]^T.
    \end{align}
    \item $\bm{J}_{C,i}$ and $\bm{f}_{C,i}$ are the Jacobian matrix and contact force at the contact point with the external environment.
\end{itemize}

$\bm{\tau}_R$ is 
the joint torque vector in the skeletal system.
We considered only the passive internal forces $\bm{\tau}_S$ caused by the viscoelasticity of the prosthesis, as follows:
\begin{align}
    \bm{\tau}_\mathrm{S} = \bm{K}(\bm{q}_\mathrm{S, eq} - \bm{q}_\mathrm{S}) - \bm{D}\dot{\bm{q}}_\mathrm{S},
\end{align}
where $\bm{q}_\mathrm{S, eq}$ is the value of $\bm{q}_{\rm S}$ when the prosthesis is unloaded, and
$\bm{K}$ and $\bm{D}$ are the stiffness and viscosity matrices, respectively.
These material parameters of the prosthesis were estimated
based on running and walking measurement data, using the method proposed in \cite{shimane2022, shimane2024}.

Consequently, the skeletal motion of unilateral transtibial amputee was modeled with 33 DOFs: 9 DOFs for joint flexion–extension of the upper and lower limbs; 18 DOFs for prosthetic strain; and 6 DOFs for the base link. 
\addedRone{The skeletal model itself was three-dimensional but simplified such that only pitch-rotational joints were retained, which constituted the minimum configuration required for forward locomotion.} 
In principle, $\bm{\tau}_0 = \bm{0}$; however, a virtual feedback force was applied to maintain the pelvis posture within the sagittal plane.

\addedRone{The contact force between a foot and the ground was calculated using perfectly inelastic collision model compensated by a spring-damper model \cite{ishigaki2025_}. The spring coefficient was $8.0$ kN/m, and the damping coefficient was defined as Rayleigh damping.
Based on the methodology reported in the literature \cite{ishigaki2025_}, an implicit numerical integration method was employed for FD simulation of the hybrid-link system.
The time step of the numerical integration was set to $2\times10^{-4}$ s.}

\subsection{Motion Capture Measurement\label{sect:mocap_measurement}}
We measured the walking motion of a 15-year-old male transtibial amputee (175 cm, 53 kg) using a sports prosthesis (Runner 1E91, \addedRone{SPR-3 category}), as shown in Fig. \ref{fig:model}.
The participant underwent a right transtibial amputation, with approximately one-third of the tibia remaining.
The participant provided informed consent for the protocol, which was approved by the institutional review board before the experiment. Ethics approval for the experimental procedures was granted by the Human Research Committee at the Graduate School of Information Science and Technology, the University of Tokyo (approval number: UT-IST-RE-220209).
Optical motion capture (Eagle-4, Raptor-4, Motion analysis\addedRone{, USA}) recorded walking and running motions, whereas force plates (Kistler\addedRone{, Switzerland}) measured GRFs at sampling rates of 200 Hz and 1 kHz, respectively.
\replacedRone{The markers were attached to the subject according to the Helen Hayes marker set, with 31 markers placed on the body. Three markers were attached to the socket and 21 to the prosthesis, as shown in Fig.~\ref{fig:model} (a) and (c).}{
In addition, 31 retroreflective markers were attached to the body, 3 to the socket, and 21 to the prosthesis, as shown in Fig. \ref{fig:model} (a), (c). 
The markers attached to the prosthesis were placed at the segment boundary positions of the PCS model.
}

\addedRone{The geometrical and inertial parameters of the skeletal model, including link length and mass, were scaled using subject anthropometric information.}
The generalized coordinates $\bm{q}$ were obtained by solving the inverse kinematics \cite{shimane2024} using \replacedRone{motion}{motin} capture data.
The resulting joint trajectories served as reference data for imitation learning, reported in Section \ref{section:reward}.
\begin{figure*}[t]
\centering

\begin{minipage}{0.45\hsize}
  \centering
  \includegraphics[width=\linewidth]{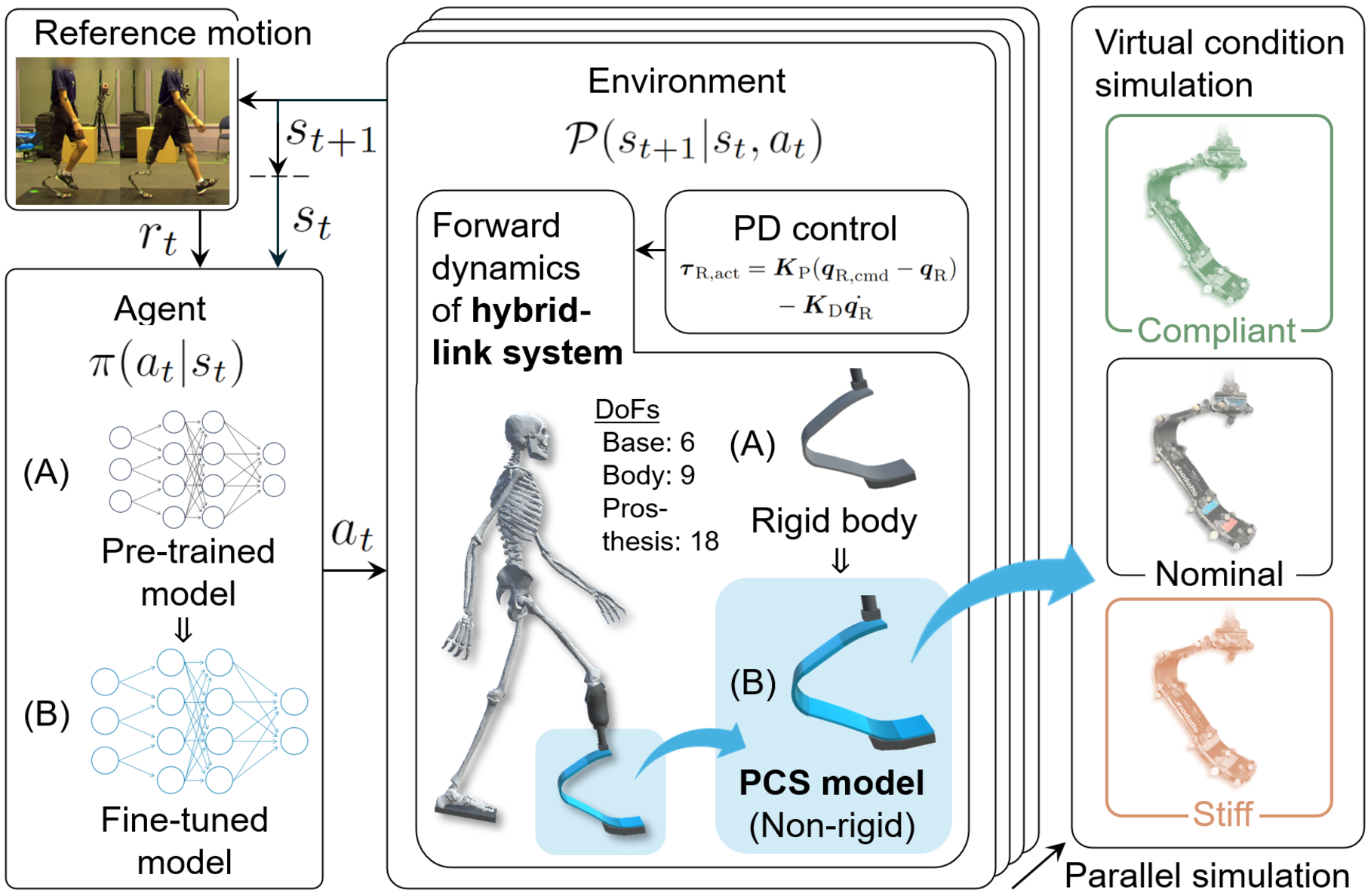}
  \captionof{figure}{Framework of reinforcement learning with the hybrid-link system. (A) Rigid-link system of the prosthesis used in pre-training, (B) Hybrid-link system.}
    \vspace{-1.5em}
  \label{fig:framework}
\end{minipage}
\hfill
\begin{minipage}{0.53\hsize}
  \centering
  \includegraphics[width=\linewidth]{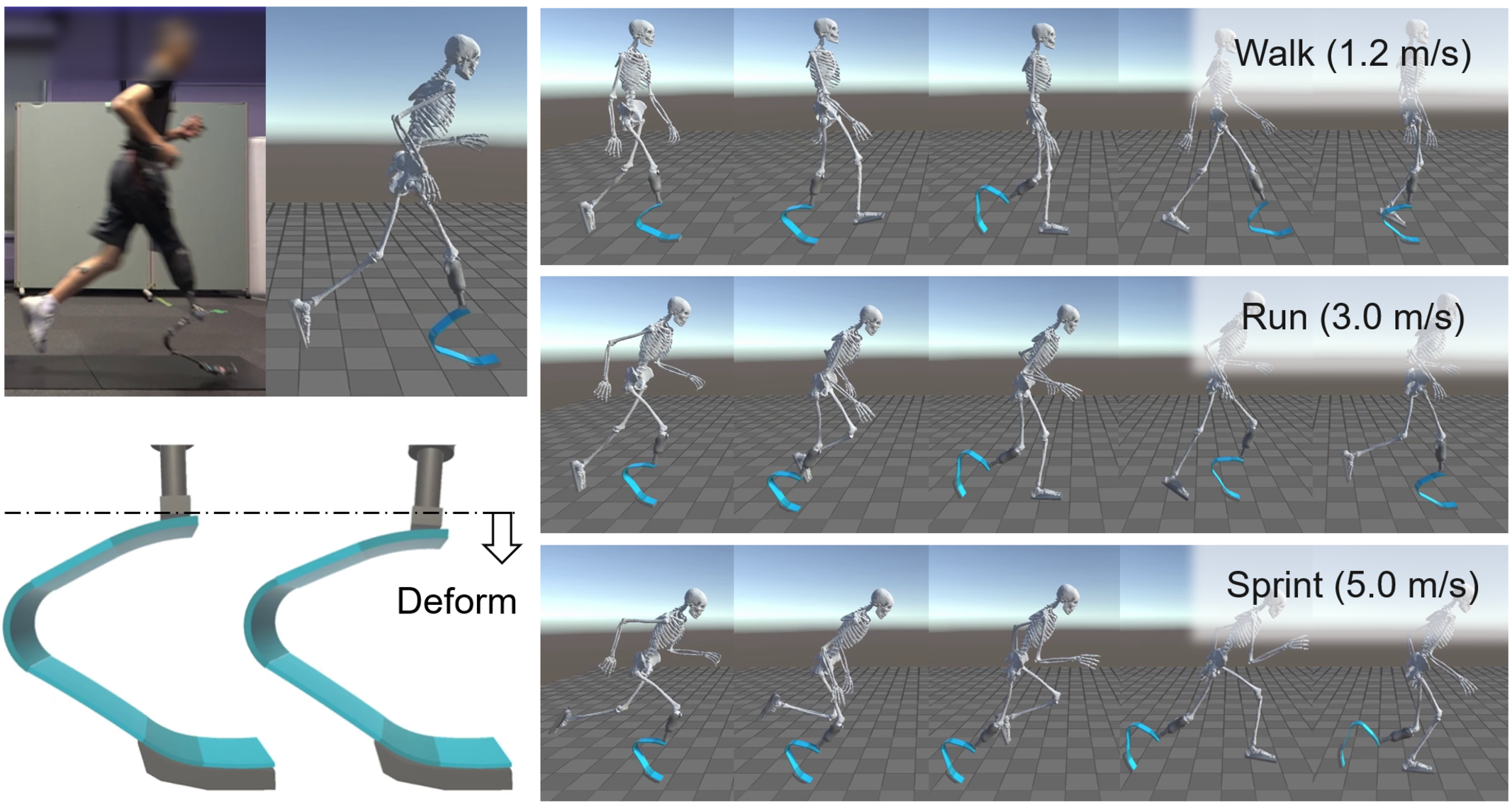}
  \captionof{figure}{Forward dynamics simulation of prosthetic motions: using a trained control policy, whole-body motions that utilize \replacedRone{prosthetic}{proshtetic} flexibility were reproduced by combining the hybrid-link system and reinforcement learning: walking, running, and sprinting.}
    \vspace{-1em}
  \label{fig:simulation}
    \vspace{-1.0em}
\end{minipage}

\end{figure*}

\section{Reinforcement Learning}
\label{sec:RL}
\subsection{Policy Gradient Method}
Reinforcement learning (RL) models an agent interacting with an environment: at each step $t$ the agent observes a state $s_t$, takes an action $a_t$, receives a reward $r_t$, and transitions to the next state $s_{t+1}$, as shown in Fig. \ref{fig:framework}. The policy $\pi_{\theta}(a_t|s_t)$ parameterized by $\theta$ is learned by maximizing the cumulative reward 
$   J(\theta) = 
    \mathbb{E}_{\pi_\theta}
    \left[
        \sum_{t=0}^{\infty}\gamma^t r_{t}
    \right],
$
where $\gamma \in [0, 1]$ is the discount rate, and $r_t$ is the reward.

The policy $\pi_{\theta}(a_t | s_t)$ is modeled as a Gaussian distribution, which
is implemented as a fully connected MLPs with two hidden layers of 64 units, and Proximal Policy Optimization (PPO) \cite{Schulman2017} with clipping to stabilize updates.

\subsection{State and Action}
With reference to \cite{peng2018, Han2022}, we define the state as 
\begin{align}
    s_t = \{ \bm{H}_0, \bm{\eta}_0, \bm{q}_{\rm R}, \dot{\bm{q}}_{\rm R} \}.
\end{align}
Note that, similar to the process in which human's neural system generate motor commands, the deformation of the prosthesis is not directly incorporated into the feedback control.  
Therefore, the strain of the prosthesis $\bm{q}_{\rm S}$ is not explicitly considered in the state representation.  
This definition is also applicable to the simulations described in the following section, where the stiffness of the prosthesis is varied.

\replacedRone{The action was defined as the reference joint angle for PD control because \cite{peng2017b} the PD control improves learning speed for specific locomotion tasks such as walking and running, compared to torque-based control. Therefore,}
{
reinforcement learning–based robot control methods use either joint torques or target values for proportional–derivative (PD) control at each joint as actions.
In this study
}
we suppose that a PD control is applied in the rigid-links as follows:
\begin{align}
    \bm{\tau}_\mathrm{R, act} = \bm{K}_\mathrm{P}(\bm{q}_{\rm R, cmd} - \bm{q}_\mathrm{R}) - \bm{K}_\mathrm{D} \dot{\bm{q}}_\mathrm{R},
\end{align}
where $\bm{K}_\mathrm{P}$ and $\bm{K}_\mathrm{D}$ are the gain matrices for PD control.
We calculate the joint torque from obtaining the command joint angle $\bm{q}_{\rm R, cmd}$ as the action $a_t$ via reinforcement learning.
We set the PD gain matrices as $\bm{K}_\mathrm{P} = 100\bm{E}$ and $\bm{K}_\mathrm{D} =  \bm{E}$.
Note that the frequency of PD control and the policy update frequency are set to 30 Hz, which is based on the firing frequency of human motor neurons.
\begin{table}
\renewcommand{\arraystretch}{1.1}
\caption{Values of Weight and Coefficient}
\label{table:weight}
\vspace{-0.5em}
\begin{tabular}{llllllllll}  
\toprule 
      & $w_q$  & $w_v$ & $w_e$ & $w_0$ & $\beta_{\rm q}$ & $\beta_{\rm v}$ & $\beta_{\rm e}$ & $\beta_{0, 1}$ & $\beta_{0, 2}$ \\ 
\midrule
Value & 1.0   & 1.0  & 1.0  & 0.1  & 2.0  & 0.03  & 40.0  & 10.0  & 0.1 \\ 
\toprule
\end{tabular}
\renewcommand{\arraystretch}{1.0}
  \vspace{-2.0em}
\end{table}

\subsection{Reward}
\label{section:reward}
Imitation learning trains the policy to imitate the expert's data, similar to supervised learning under behavior cloning.
Imitation learning is useful for efficiently learning tasks where it is difficult to design rewards. In this study, we use imitation learning based on the data obtained from the motion capture as a reference value.
Referring to previous studies \cite{peng2018},  the reward function $r_t$ is defined as follows:
\begin{align}
    r_t(s_t, a_t) =
    \sum_{i= \{ q, v, e, 0\}} w_i r_i,
    \label{eq:reward}
\end{align}
where $r_i$ and $w_i$ are the reward coefficient and its weight coefficient, respectively, as shown in Table \ref{table:weight}.
For notational convenience, we define a function $r(\beta, \bm{x}_{\rm ref}, \bm{x})$ as follows:
\begin{align}
    r(\beta, \bm{x}_{\rm ref}, \bm{x})
    =
    \exp (-\beta \| \bm{x}_{\rm ref} - \bm{x}\|^2).
\end{align}

The subscript ''ref'' represents the reference value calculated on the basis of the motion capture measurement data.
Then, each reward term is defined as below.
\begin{itemize}
    \item 
    $r_\mathrm{q} := r(\beta_{\rm q}, \bm{q}_{\rm R, ref}, \bm{q}_{\rm R})$ and $r_\mathrm{v} := r(\beta_{\rm v}, \dot{\bm{q}}_{\rm R, ref}, \dot{\bm{q}}_{\rm R})$ are mimicking-terms of the joint angle $\bm{q}_{\mathrm{R}}$ and its velocity.
    \item 
    $r_\mathrm{e}:= r(\beta_{\rm e}, \bm{p}_{ref}, \bm{p})$ is a mimicking-term for
    the positions of the end effectors $\bm{p} = [\bm{p}_{\rm head}^T, \ \bm{p}_{\rm r,hand}^T, \ \bm{p}_{\rm l,hand}^T, \\ \bm{p}_{\rm r,foot}^T, \ \bm{p}_{\rm l,foot}^T]^T$ w.r.t. the world frame.
    \item 
    $r_0 := r(\beta_{0,1}, \bm{q}_{0, \mathrm{ref}}, \bm{q}_0) +  r(\beta_{0,2}, \dot{\bm{q}}_{0, \mathrm{ref}}, \dot{\bm{q}}_0)$ is a mimicking-term for the base link position and velocity where $\bm{q}_{0}$ is a 6D vector representation of $\bm{H}_{0}$.
\end{itemize}

\section{Prosthetic \replacedRone{Running}{Walking} by Imitation Learning}
\label{sec:imitation}
\subsection{Implementation of Reinforcement Learning}
\subsubsection{Fine-tuning}
A fast dynamic computation method using implicit integration has been proposed for the hybrid-link system \cite{ishigaki2025_}.
However, simulations with flexible deformation are much more computationally expensive than rigid-link systems, making it essential to improve learning efficiency and reduce simulation time over tens of thousands of iterations.
As shown in Fig.~\ref{fig:framework}, we first train a policy 
using a rigid-link model instead of the PCS model, and
then fine-tuned it with the hybrid-link system to account for prosthesis deformation.
This 
process provides effective initial parameters, enabling efficient learning with fewer iterations.

\subsubsection{Multiprocessing}
To improve computational efficiency, interactions are simulated across multiple independent CPU sub processes to collect empirical data, and the policy is updated on the main process using the aggregated results.  
The time-series empirical data for one episode in the $j$th ($j = 1, 2, \dots, n$) process, $\Gamma_j = \{\bm{s}_j, \bm{a}_j, \bm{r}_j\}$, are aggregated into a single buffer data $\Gamma = \{\Gamma_1, \Gamma_2, \dots, \Gamma_n\}$ through inter-process communication.
Here $\bm{s}_j = \{s_1, s_2, \dots, s_t, \dots\}$, $\bm{a}_j = \{a_1, a_2, \dots, a_t, \dots\}$, and $\bm{r}_j = \{r_1, r_2, \dots, r_t, \dots\}$ are the states, actions, and rewards of the hybrid-link system at each episode in the $j$th process.
The neural network parameters of the policy, including weights and biases, are then updated based on the aggregated empirical data.

\subsubsection{Implementation and Parameter Settings}
We used PyTorch and Stable-Baselines3 to implement RL.
As the FD calculation of the hybrid-link system is implemented in C++ for robot control, we connected this library and frameworks using Pybind11.
We set the hyper parameters for learning as follows: buffer size = 4096, mini-batch size = 256, learning rate = 0.0003, discount rate ($\gamma$) = 0.95, and GAE coefficient ($\lambda$) = 0.95. 
The calculations for learning are performed using 8 processes for the rigid-link system and 16 processes for the hybrid-link system on the CPU.

\begin{figure}[t]
  \centering
    \includegraphics[width=0.9\hsize]{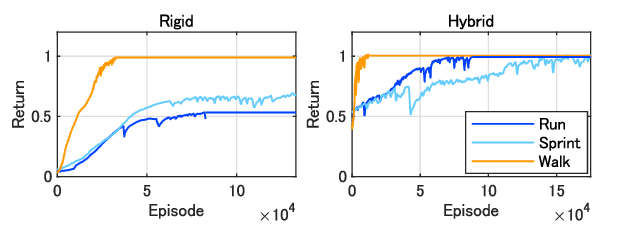}
  \vspace{-0.5em}
  \caption{Learning return on prosthetic \addedRone{walking, }running\addedRone{, sprinting}: (Left) Return in the pre-training using a rigid-link system. (Right) Return in fine-tuning using the hybrid-link system that considers the prosthesis's deformation. }
  \label{fig:return}
  \vspace{-1em}
\end{figure}
\begin{figure}[t]
  \centering
    \includegraphics[width=0.8\hsize]{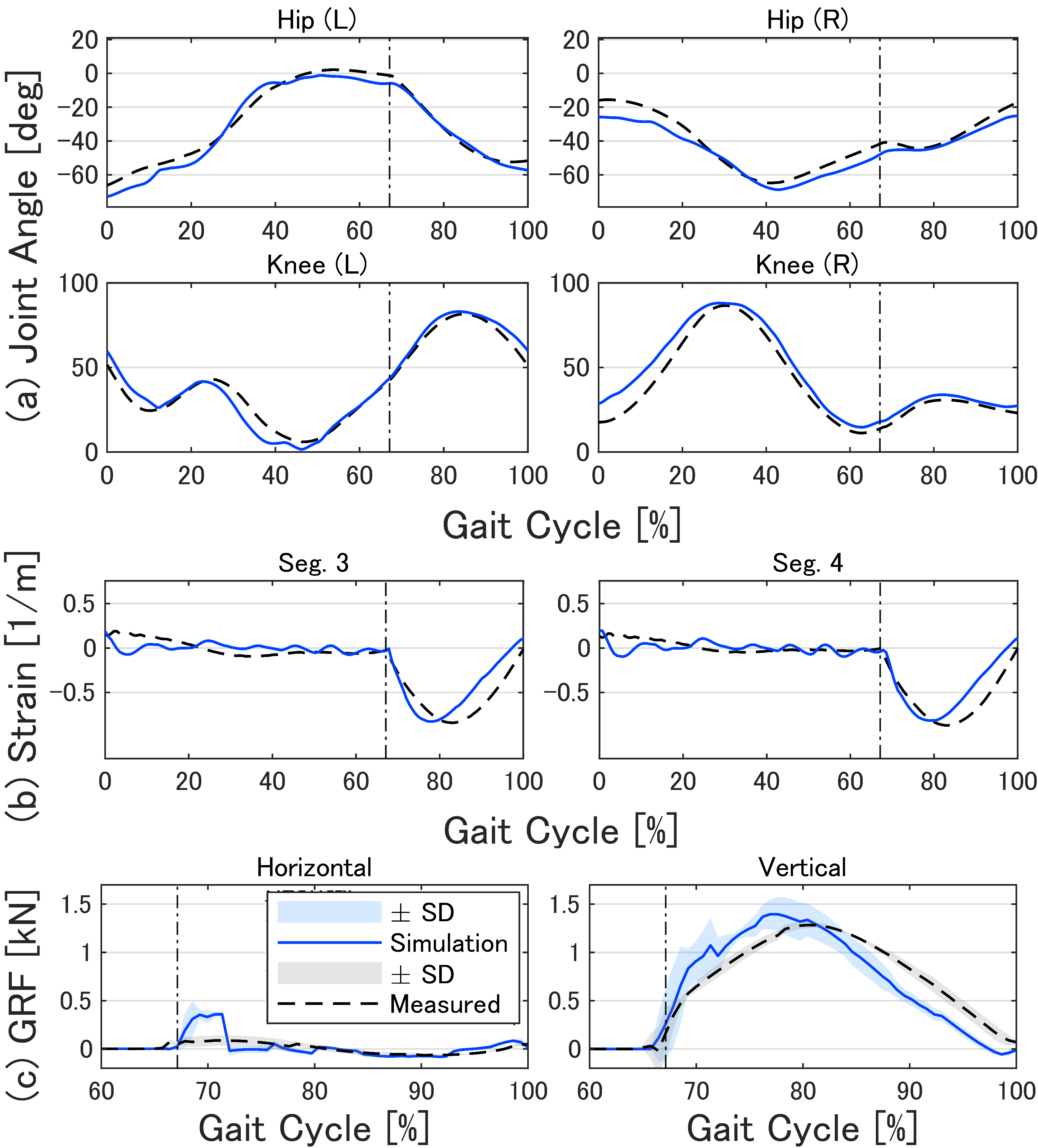}
  \caption{Running simulation results: (a) \addedRone{mean angles of hip and knee joints, Left: intact side)}, (b) \addedRone{mean} prosthetic strain \deletedRone{of the 3rd and 4th segments }of the PCS model, (c) \addedRone{mean} GRFs in the horizontal: braking and vertical: body-support directions \replacedRone{with shaded bands indicating the standard deviations.}{The blue line shows the simulation results using the trained controller, and the dashed black line shows the measured values. }}
  \label{fig:reslut_joint_strain_grf}
  \vspace{-1.5em}
\end{figure}
\begin{table}[t]
\footnotesize
\centering
\setlength{\tabcolsep}{3pt}
\caption{\addedRone{RMSE and peak CI half-width for joint angles, strain, and GRF.}}
\label{table:rmse_ci}
\vspace{-0.5em}
\begin{tabular}{lcccccc}
\toprule

\makecell{Joint angle \\ RMSE [\%] \\ CI [deg]}
& \makecell{Hip (L) \\ 6.0 \\ 1.84}
& \makecell{Hip (R) \\ 7.8 \\ 2.10}
& \makecell{Knee (L) \\ 7.1 \\ 2.16}
& \makecell{Knee (R) \\ 9.5 \\ 2.63}
& \makecell{Ankle (L) \\ 13.7 \\ 2.00}
& \makecell{All \\ 9.2} \\

\midrule

\makecell{Strain (Stance) \\ RMSE [\%] \\ CI [1/m]}
& \makecell{Seg.1 \\ 8.0 \\ 0.031}
& \makecell{Seg.2 \\ 6.1 \\ 0.068}
& \makecell{Seg.3 \\ 22.8 \\ 0.118}
& \makecell{Seg.4 \\ 22.3 \\ 0.103}
& \makecell{Seg.5 \\ 11.6 \\ 0.030}
& \makecell{All \\ 16.1} \\

\midrule

\makecell{Force (Stance) \\ RMSE [\%] \\ CI [kN]}
& \makecell{Fx \\ 7.5 \\ 0.04}
& \makecell{Fy \\ 4.5 \\ 0.05}
& \makecell{Fz \\ 16.7 \\ 0.25}
& 
& 
& \makecell{All \\ 10.9} \\

\bottomrule
\end{tabular}
  \vspace{-2.5em}
\end{table}

\subsection{Result}
Fig. \ref{fig:simulation} illustrates the walking (1.2 m/s), running (3.0 m/s) and sprinting (5.0 m/s) simulations of the hybrid-link system controlled by the trained policy.
Fig.~\ref{fig:return} presents the cumulative reward per training episode for running; pre-training converged after 82,000 episodes (33 h), whereas the hybrid-link system required 88,000 episodes (102.5 h).  

Fig.~\ref{fig:reslut_joint_strain_grf} compares the simulation results with the experimental data: (a) right hip and knee joint angles \addedRone{(Right: prosthetic side, Left: intact side)}, (b) prosthetic strain in the 3rd and 4th PCS segments, and (c) GRF in the horizontal (braking) and vertical (support) directions.  
\addedRone{The plotted data correspond to one gait cycle starting from the prosthetic right-foot toe-off.}
\addedRone{The blue solid lines and shaded bands indicate the mean and standard deviation of trajectories obtained from 5 episodes of the simulation, in which each episode was generated from action sampled from the trained policy.}
\addedRone{Table~\ref{table:rmse_ci} shows peak confidence interval (CI) half-width for these episodes, and shows the root mean square errors (RMSEs) between the simulated and measured trajectories, expressed as the ratio from the corresponding value range.}

The dotted line marks the transition from the swing phase to the stance phase. 
In Fig.~\ref{fig:reslut_joint_strain_grf} (b) during the right-leg stance phase, the prosthesis deforms from ground contact to lift-off because of the body dynamics and viscoelasticity, resulting in strain.
Although prosthetic strain is not explicitly included in the reward and state, both the strain and GRF are similar to the reference data.
\deletedRone{These results suggest that passive prosthesis deformation is implicitly captured in the training process, allowing the control policy to adapt to its flexibility.\\
Fig.~6 compares the return between the walking and running simulations.
Notably, in the walking simulation, the final return values were identical regardless of whether the prosthesis was modeled as a rigid or flexible body, whereas these values differed in the running simulation. 
This indicates that, unlike walking, a rigid-body prosthetic model cannot accurately reproduce a subject's running motion. 
This difference arises from the distinct GRF characteristics during the contact phase. 
As summarized in Table~\ref{table:GRF}, compared with the rigid-link, the hybrid-link presented approximately twice the propulsive force in the latter half of the right stance phase, a 23\% longer contact time, and twice the time to reach the peak GRF.
These results indicate that the running motion of an amputee wearing a leaf spring-type prosthesis can be reproduced only by utilizing the flexibility of the prosthesis, and that the proposed hybrid-link-based method enables us to obtain such dynamic motion controllers.}

\section{Effect of Prosthetic Stiffness Changes}
\label{sec:effect}
\subsection{Running Simulation with Different Stiffness Conditions}
Once FD simulations of prosthetic running become possible, it will be possible to clarify how mechanical properties of the prosthesis affect movement.
We demonstrate that running motion with virtually varied stiffness can be realized through FD simulations based on the proposed learning framework, as shown in Fig. \ref{fig:framework}. 
Beck et al. \cite{beck2017} experimentally investigated the effects of prosthetic stiffness on running performance when the stiffness level was set to one category above or below  the recommended stiffness based on body weight, 
\addedRone{corresponding to approximately $-10\%$ and $+10\%$ differences, respectively.}
On the basis of the previous study, we define the set of prosthetic stiffness conditions as follows.
\begin{enumerate}
    \item {\bf Nominal} condition\replacedRone{ is used by subject for the motion measurement in Section \ref{sect:mocap_measurement}. The stiffness was selected based on the subject's body weight and corresponds to the SPR-3 category of the 1E91.}{uses the original value}
    \item {\bf Compliant} condition uses the stiffness value that is $-10\%$ of the Nominal value. 
    \item {\bf Stiff} condition uses the stiffness value that is $+10\%$ of the Nominal value.
\end{enumerate}

\replacedRone{The policy was trained under the {\it Nominal} condition in Section \ref{sec:imitation}, and then fine-tuned for {\it Compliant} and {\it Stiff} conditions using the same joint trajectories.
All three policies were trained until the return values in (\ref{eq:reward}) became equal.}{
The policies are trained using these three stiffness values to replicate the reference motion until the return values in (\ref{eq:reward}) become equal.}

\begin{figure}[t]
  \centering
    \includegraphics[width=0.9\hsize]{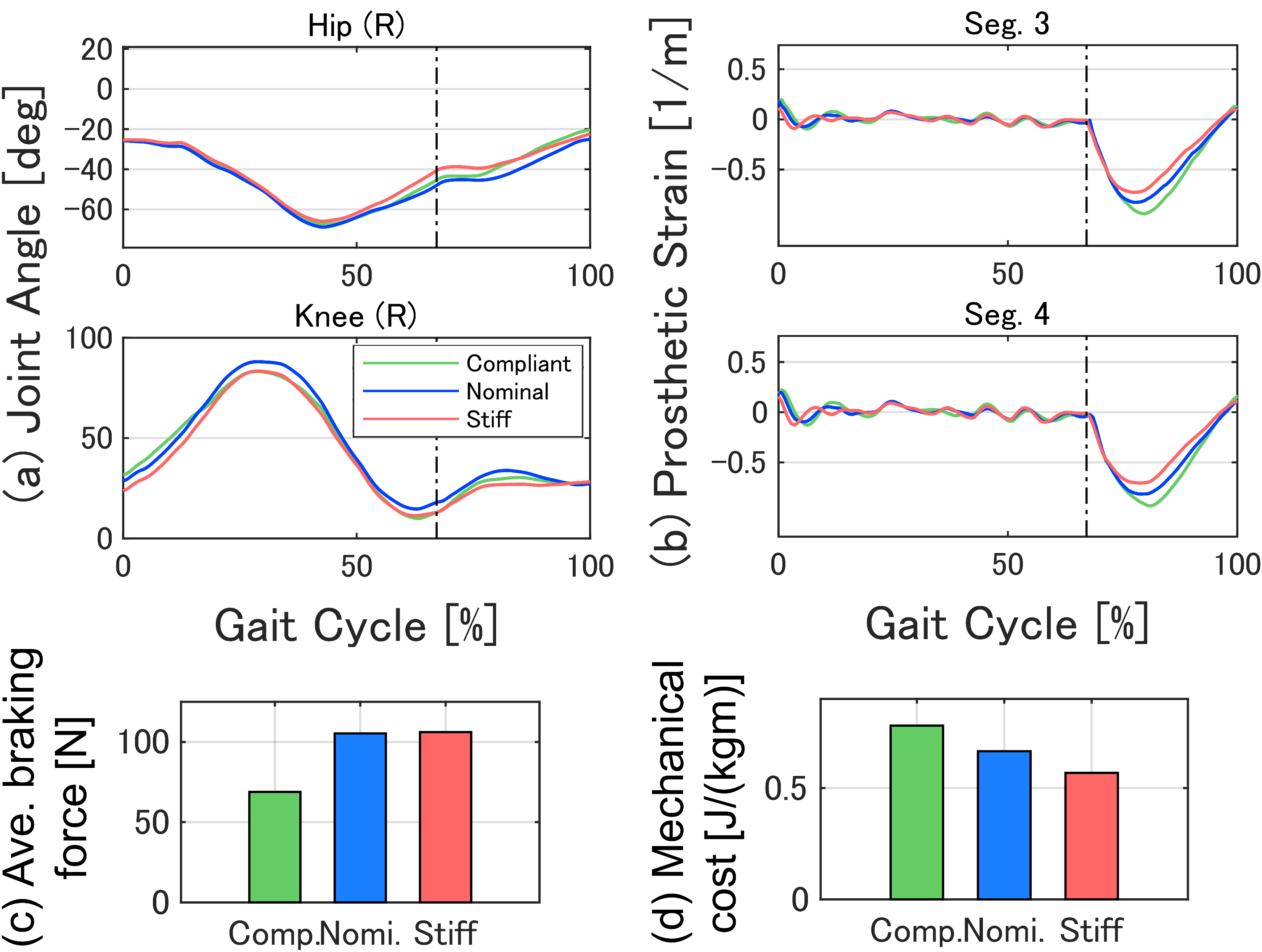}
    \vspace{0.5em}
  \caption{
  Simulation results under different stiffness conditions: (a) angle of \deletedRone{the hip and knee }joints, (b) prosthetic strain of \deletedRone{the 3rd and 4th segments of }the PCS model, (c) average GRF\deletedRone{in the horizontal braking direction during the late stance phase}, and (d) mechanical cost\deletedRone{, which is the elastic energy per body weight and distance at the moment of maximum prosthetic deformation. The {\it Nominal}, {\it Compliant}, and {\it Stiff} conditions are represented by green, blue, and red lines and bars, respectively}.}
  \label{fig:changestiff}
  \vspace{-1em}
\end{figure}
\begin{figure}[t]
  \centering
    \includegraphics[width=0.6\hsize]{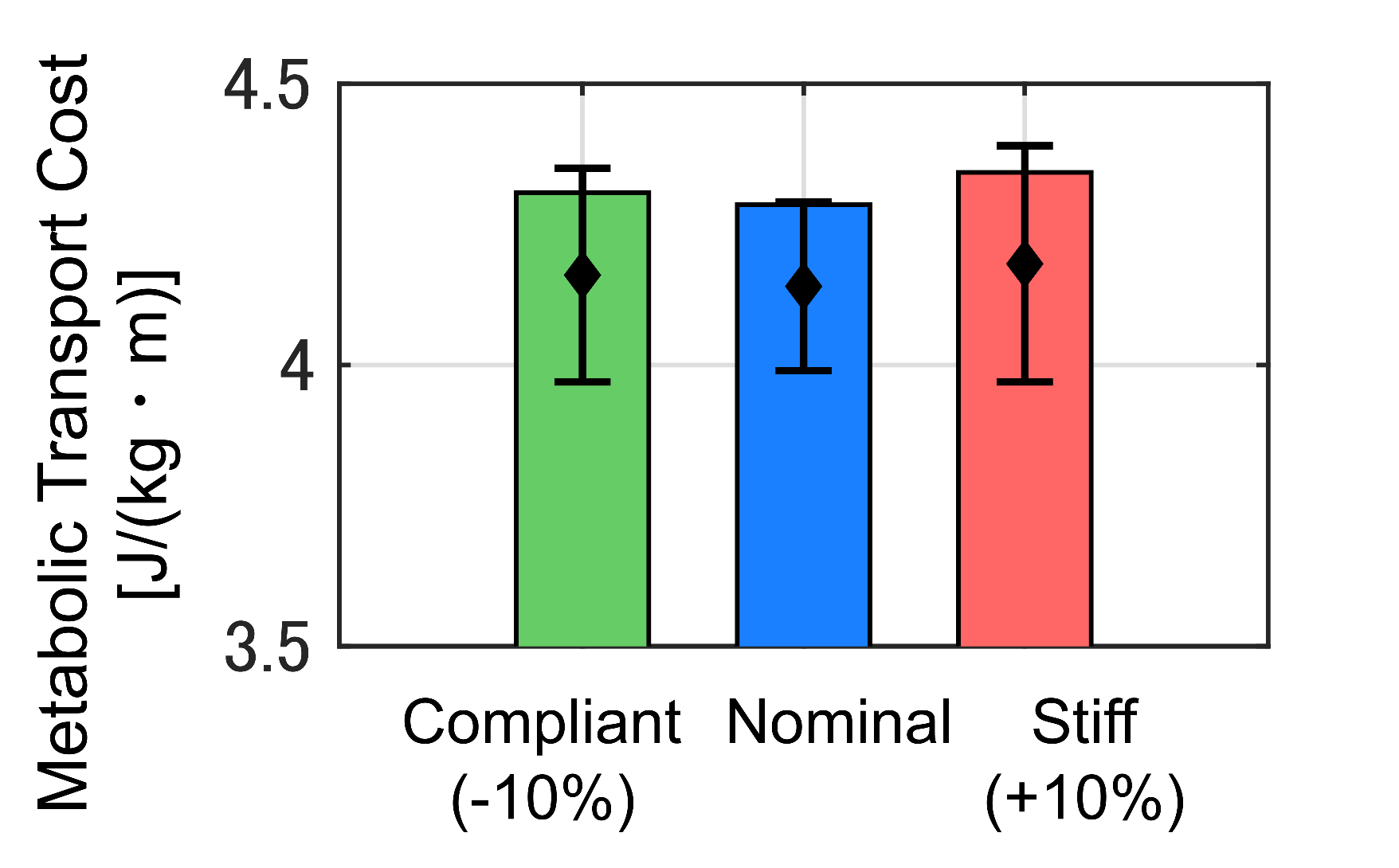}
  \caption{
  Metabolic costs of transport obtained from running simulations with different prosthetic stiffnesses are shown, with error bars indicating values reported in \cite{beck2017}.}
  \label{fig:energy_cost}
  \vspace{-1.5em}
\end{figure}

\subsection{Results}
\subsubsection{Effect of Prosthetic Stiffness on Simulated Running Motion}
Fig. \ref{fig:changestiff} (a) presents the simulation results of the angles of the right hip and knee joints, (b) the strains of the 3rd and 4th segments of the prosthesis, (c) the average GRF in the horizontal braking direction during the late stance phase, and (d) the mechanical cost, which is the elastic energy per body weight and distance at the moment of maximum prosthetic deformation. 
In Fig. \ref{fig:changestiff} (b), after ground contact, a lower stiffness resulted in greater deformation. The prosthesis under the {\it Compliant} condition exhibited up to 32\% greater strain than the prosthesis under the {\it Stiff} condition. 
In Fig. \ref{fig:changestiff} (c), the prosthesis under the {\it Compliant} condition presented the lowest GRF, which was 35\% lower than that under the {\it Stiff} condition. 

\subsubsection{Effect of Prosthetic Stiffness on Energy Consumption}
If the dynamics of body motion, such as joint power, can be computed, energy consumption during running can be simulated.
Among the three conditions, the CoT $E_{\rm CoT}$, the energy to move 1 m per 1 kg of body mass, are calculated from the joint mechanical work as follows:
\begin{align}
\label{eq:mechanical_work}
    E_{\rm CoT} = \frac{1}{\alpha_{e}}\frac{E_{\tau}}{md},
    \
    E_{\tau} = \int_0^T \sum_{j=1}^n \max\left(0, \tau_j\dot{\theta}_j\right)\,dt
\end{align}
where $E_{\tau}$ represents the positive mechanical work performed by all joints, and $n$ is the number of joints. $\alpha_{e} = 0.8$ is the energy transfer efficiency \cite{riddick2022}, $m$ is the total body mass including the prosthesis, and $d$ is the distance traveled during the simulation time $T$.

Fig. \ref{fig:energy_cost} shows the CoT $E_{\rm CoT}$ under the three stiffness conditions. 
Under the {\it Nominal} condition, the running CoT is 4.3 J/kg·m, which is the smallest value among the three conditions.
The black line indicates the mean and standard error range of the CoT reported in \cite{beck2017}, which was measured in unilateral transtibial amputees using gas analysis under the same running speed conditions.
The simulated values under all three conditions fall within this reported range.  
\begin{table}[t]
\centering
\renewcommand{\arraystretch}{1.1}
\caption{Comparison of GRF between the Rigid and Hybrid-links on Running Simulation}
\label{table:GRF}
\vspace{-0.5em}
\begin{tabular}{lcc}
\toprule
 & Rigid-body & Hybrid-link \\
\midrule
Average propulsion force [N] & 92.9 & 187.5 \\
Contact time [s] & 0.2 & 0.26 \\
Time to peak GRF [s] & 0.065 & 0.13 \\
\bottomrule
\end{tabular}
\renewcommand{\arraystretch}{1.0}
  \vspace{-1em}
\end{table}

\deletedRone{These simulation results showed that stiffness variation affects prosthetic deformation, joint trajectories and the GRF.
As shown in Fig.~\ref{fig:changestiff}, increased flexibility reduced impact at landing, and the learned policies adapted successfully to these changes.
This adaptability is obvious in the consistent joint trajectories shown in Fig.~\ref{fig:changestiff} (a) to reproduce reference running motion despite the variation in prosthetic stiffness.\\
As shown in Fig.~\ref{fig:energy_cost}, under the {\it Stiff} condition, the horizontal braking force at ground contact increased, requiring more propulsive effort for body acceleration, resulting in a higher cost.  
In contrast, under the {\it Compliant} condition, the elastic energy stored in the prosthesis increases, as shown in Fig. \ref{fig:changestiff} (d), and since this energy is supplied by the body's mechanical work, the overall cost also increases.
One possible reason for this is that additional force is required for posture control and propulsion when using a compliant prosthesis.
These findings indicate that an appropriate stiffness, neither too soft nor too stiff, relative to body weight and running motion can suppress excessive energy consumption. 
Overall, the results show that the proposed method can adaptively simulate realistic running motions while reflecting stiffness-dependent variations and maintaining biomechanical plausibility.\\
In this study, as a first step toward realizing motion simulation of a unilateral transtibial amputee, walking and running motion data from a single subject were used for imitation learning and evaluation.
Therefore, the simulated motions are limited to reproducing an individual's movements including their specific body size and characteristics.
In addition, the simulation results under virtual prosthetic stiffness variations were validated only through comparisons with the results of previous study.
When simulating the effects of large differences from the actual stiffness, further validation comparing the estimated results with experimental measurements is needed.}

\section{\addedRone{Discussion}}
\label{sec:discussion}
\subsection{\addedRone{Running Simulation with RL}}
In the walking simulation, as reported in Fig.~\ref{fig:return}, the final return values were identical regardless of whether the prosthesis was modeled as a rigid or flexible body, whereas these values differed in the running simulation. 
This indicates that, unlike walking, a rigid-body prosthetic model cannot accurately reproduce a subject's running motion. 
This difference arises from the distinct GRF characteristics during the contact phase. 
As summarized in Table~\ref{table:GRF}, compared with the rigid-body, the hybrid-link presented approximately twice the propulsive force in the latter half of the right stance phase, a 23\% longer contact time, and twice the time to reach the peak GRF.
These results 
indicate that the prosthetic running motion can be reproduced only by utilizing the prosthetic flexibility, and that the proposed hybrid-link-based framework enables the generation of such dynamic motion controllers\addedRtwo{, which directly addresses the first research question presented in Section \ref{sec:intro}}.

\addedRone{
The results shown in Fig. \ref{fig:reslut_joint_strain_grf} and Table \ref{table:rmse_ci} indicate that the FD simulation with the trained control policy reproduces motion that is similar to the reference motion, although it does not precisely track the reference trajectories.
However, precise trajectory tracking is not the primary objective of employing FD simulation in this study.
Instead, it is more important that the FD simulation exhibits robustness when simulating conditions that differ from the reference conditions, which is further examined in the following subsection.
}

\subsection{\addedRone{Applicable Range of Obtained Policy}}
\begin{figure}[t]
  \centering
    \includegraphics[width=0.8\hsize]{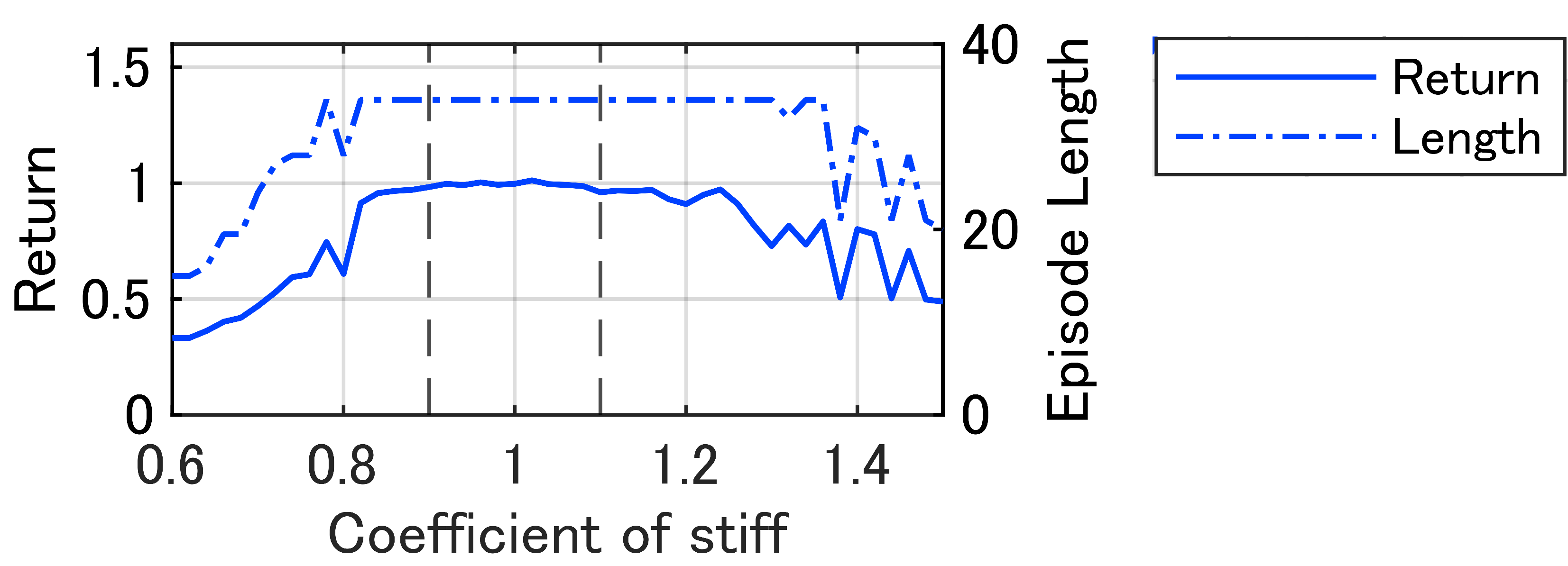}
  \caption{\addedRone{Return and episode length under variation of the stiffness parameter}
  }
  \label{fig:sensityvity}
  \vspace{-2em}
\end{figure}

\addedRone{To evaluate the robustness of the simulation framework, we investigated the range of prosthetic stiffness values over which the obtained control policy can be applied without additional fine-tuning. 
Fig.~\ref{fig:sensityvity} shows the reward values and episode lengths obtained from simulations in which the stiffness parameter is varied from $-40\%$ to $+50\%$ with 2\% interval.
As shown in the figure, both the reward values and episode lengths gradually decrease when the stiffness deviates beyond approximately $\pm$20\% from the nominal value.
According to \cite{beck2016}, the C-shaped prosthesis such as the Catapult FX6 (Freedom Innovations, USA) is designed such that a $\pm$10\% change in stiffness corresponds to a $\pm$10 kg change in the recommended user body weight.
Therefore, the result of Fig.~\ref{fig:sensityvity} implies that the obtained policy can be applied to a wide range corresponding to approximately $\pm$20 kg of body weight.
However, this range is sufficiently large that, for accurate analysis across different prosthesis categories, it may be preferable to perform additional training or fine-tuning of the policy.
Simulating the effects of larger deviations from the nominal stiffness remains an important direction for future work.}

\begin{figure*}[t]
\centering

\begin{minipage}{0.3\hsize}
  \centering
  \includegraphics[width=\linewidth]{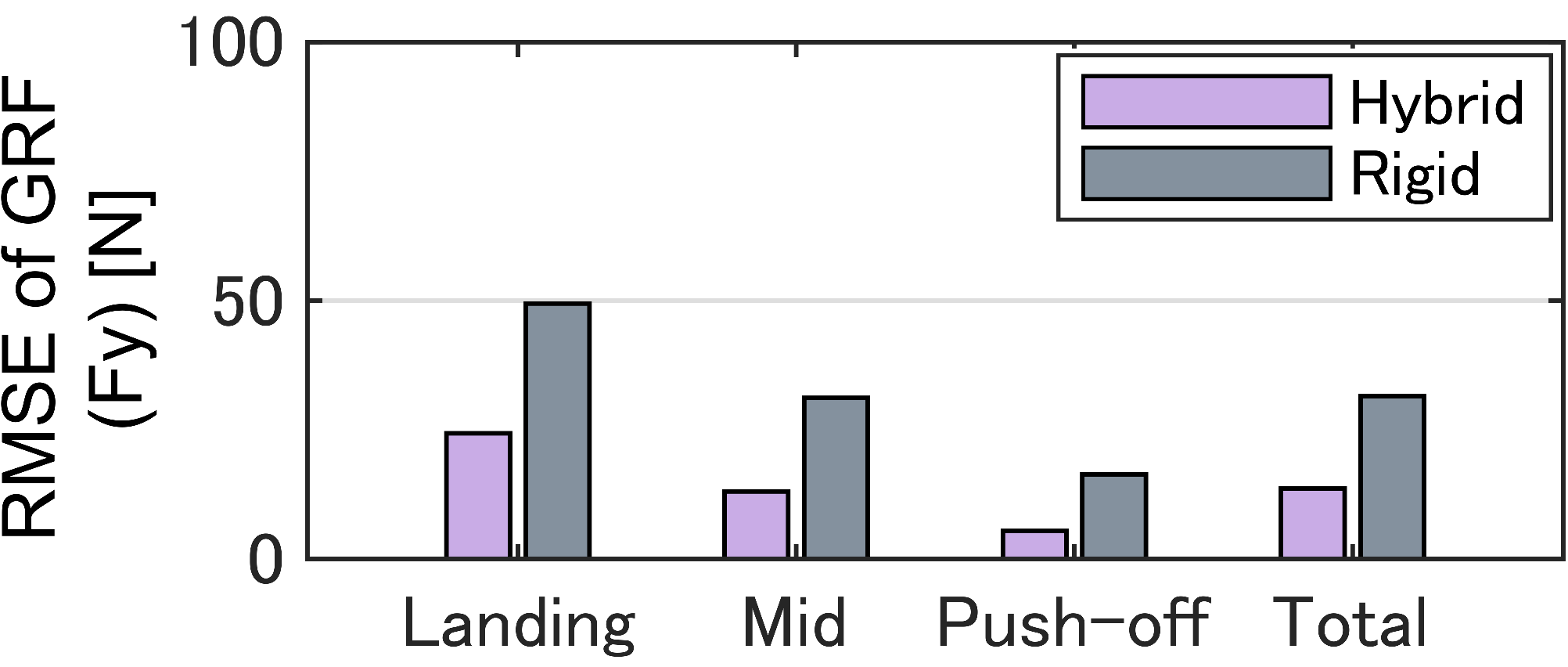}
  \captionof{figure}{\addedRone{RMSE of walking GRFs with a rigid-body multi-link model}}
  \label{fig:walk03_rigid_1}
\end{minipage}
\hfill
\begin{minipage}{0.3\hsize}
  \centering
  \includegraphics[width=\linewidth]{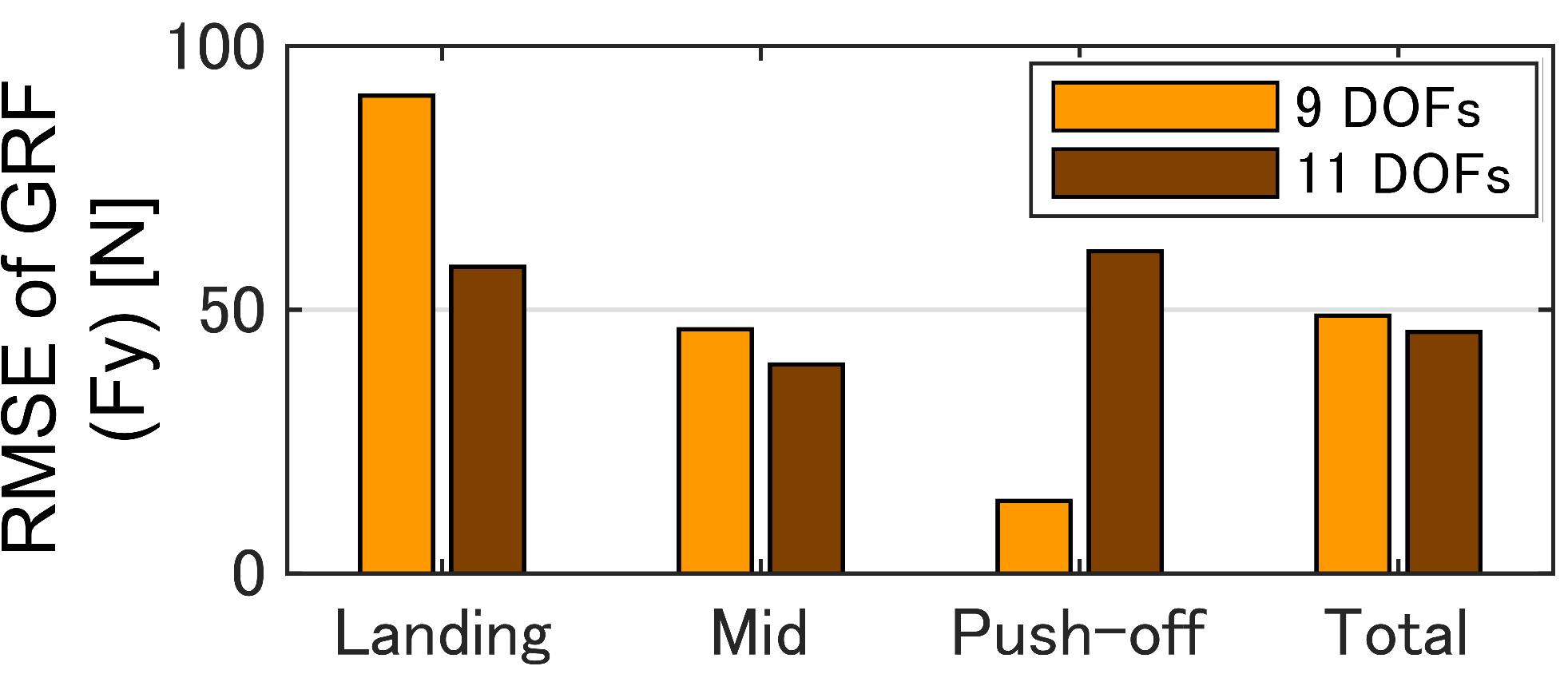}
  \captionof{figure}{\addedRone{RMSE of GRFs with additional DOF during walking}}
  \label{fig:grf_walk}
\end{minipage}
\hfill
\begin{minipage}{0.35\hsize}
  \centering
  \includegraphics[width=\linewidth]{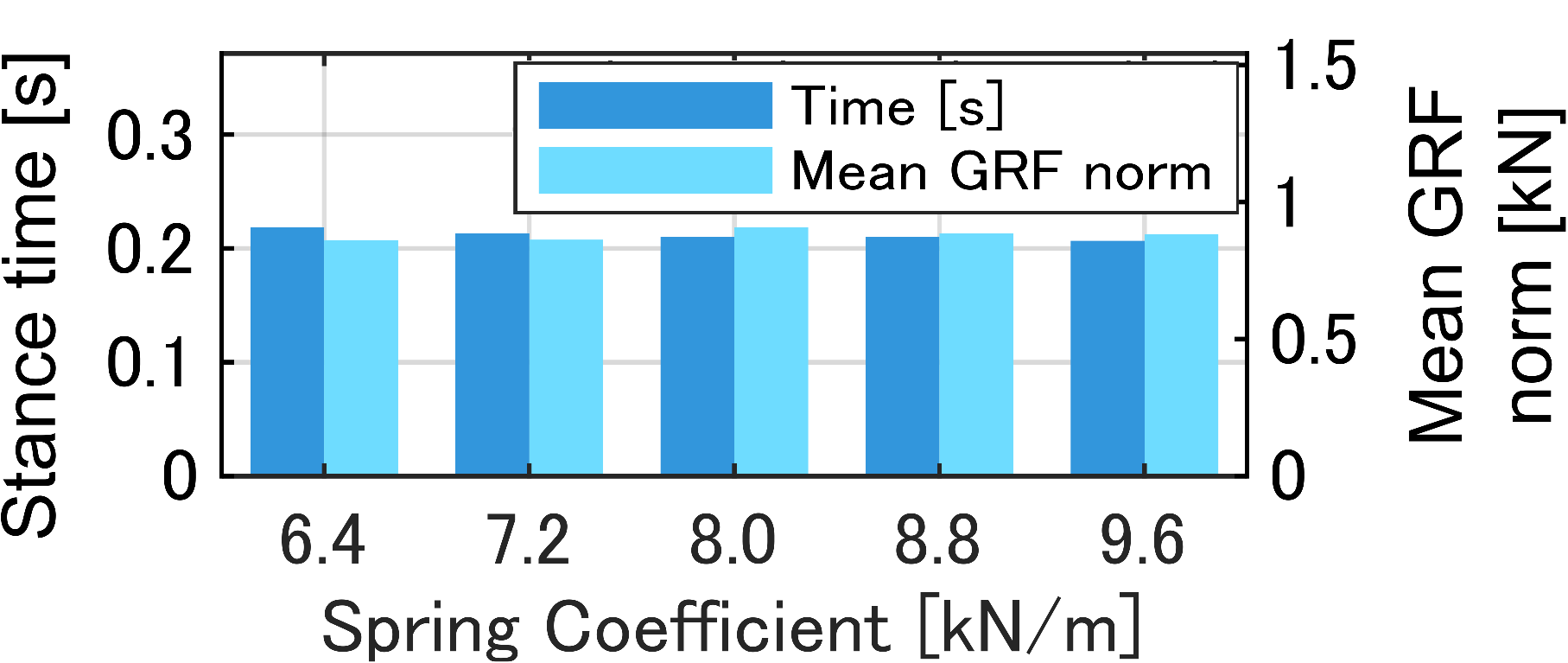}
  \vspace{-1.8em}
  \captionof{figure}{\addedRone{Stance time and mean GRF across contact parameters}}
  \label{fig:contact_time}
\end{minipage}
  \vspace{-2.0em}
\end{figure*}

\subsection{\addedRone{Simulation Results under Different Stiffness Conditions}}
As shown in Fig.~\ref{fig:changestiff}, increased flexibility reduces the impact at landing, and the learned and fine-tuned control policies adapt successfully to the different stiffness conditions.
This adaptability is evident in the consistent joint trajectories shown in Fig.~\ref{fig:changestiff} (a), which reproduce the reference running motion despite variations in prosthetic stiffness.

As shown in Fig.~\ref{fig:energy_cost}, the estimated CoT values are consistent with those reported in previous study, thereby providing validation for the second research question.
Under the {\it Stiff} condition, the horizontal braking force at ground contact increases, which requires greater propulsive effort for body acceleration and results in a higher energy cost.  
In contrast, under the {\it Compliant} condition, the elastic energy stored in the prosthesis increases, as shown in Fig. \ref{fig:changestiff} (d). Because this energy is supplied by the mechanical work of the body, the overall energy cost also increases.
One possible explanation for this observation is that additional force is required for posture control and propulsion when using a compliant prosthesis.
These findings indicate that an appropriate stiffness level, neither excessively compliant nor excessively stiff relative to body weight and running motion, can reduce excessive energy consumption.
Overall, the results demonstrate that the proposed method can simulate adaptive running motions under virtual conditions that correspond to real movement, while capturing stiffness-dependent variations and maintaining biomechanical plausibility.

\subsection{\addedRone{Limitation}}
\addedRone{A limitation is the use of joint torques to actuate the skeletal system, rather than muscle-based actuation.
In addition, \eqref{eq:mechanical_work} does not account for physiological contributions such as negative work, co-contraction, or energy transfer by multi-articular muscles. 
Consequently, $E_{\tau}$ may be overestimated, indicating the presence of a positive bias in the estimated CoT.
This bias likely contributes to higher CoT values than those reported in \cite{beck2017}, as shown in Fig. \ref{fig:energy_cost}.
Reducing this bias through extension to a muscle-actuated system is expected to yield CoT estimates closer to those reported in~\cite{beck2017}.}

\replacedRone{Another limitation is that}{Additionally,} imitation learning and evaluation are based on walking and running motion data obtained from a single subject.
The resulting simulations are therefore limited to reproducing the individual's movement, including subject-specific body size and biomechanical properties.
\addedRone{Future work will involve broader investigation across multiple subjects to improve generalizability.}

\deletedRone{Another point is that the simulation results under virtual prosthetic stiffness variations were validated only through comparisons with the results of previous study.}


\section{Conclusion}
\label{sec:conclusion}
In this study, we proposed a RL-based method for a hybrid-link system that
enables the simulation of motion in a unilateral transtibial amputee 
using 
a leaf-spring-type sports prosthesis.
The main results
are summarized as follows.
\begin{enumerate}
    \item 
    The subject's motion was modeled through FD simulation of a hybrid-link system, considering 9 DOFs for the skeletal system and 18 DOFs for the PCS model, with a virtual feedback force applied in the sagittal plane. 
    During training, the control policy was first trained using a rigid-body prosthesis model and subsequently fine-tuned using the PCS model to account for flexible deformation.
    This approach enabled efficient generation of a control policy that produces whole-body running motion utilizing prosthetic elasticity.

   \item \addedRone{Running motion was successfully reproduced in the FD simulation using the trained policy.
   The RMSEs between the simulation result and reference were 9\% for the joint trajectories and 11\% for the GRF.
   Although the trained policy does not exactly track the reference trajectories, the policy demonstrates robustness by enabling simulation without additional training under prosthesis stiffness variations of $\pm$20\%.
    }
    \item 
    Running motions were also simulated using three control policies, each trained under a different virtual stiffness condition.
    Among these conditions, the {\it Nominal} condition yields the lowest metabolic cost of 4.3 J/kg·m, indicating that moderate stiffness relative to body weight and running motion minimizes energy consumption.
    These values are consistent with those reported in previous study, thereby supporting the validity of the proposed method under virtual stiffness variations that differ from real-world conditions.
\end{enumerate}

The results indicate that future studies could investigate the effects of a wider range of prosthetic parameters, including viscosity, geometry, mounting height, and alignment angle.
Such investigations would further support the development of design and selection guidelines for assistive devices that interact dynamically with human movement.

\appendix
\subsection{\addedRone{\replacedRtwo{Simplification in}{Comparison between Rigid-link and PCS models for} Prosthesis Model\deletedRtwo{ing}}}
\addedRone{
    \replacedRtwo{We evaluate how much the GRF estimation is affected if we simplify the prosthesis model, comparing}{We compared} the GRF estimations \addedRtwo{by inverse dynamics (ID)} for the following two cases: one by \deletedRtwo{inverse dynamics (ID) of} the hybrid-link system using the PCS model, and the other by ID of a rigid\deletedRtwo{ multi}-link model in which the prosthesis leg consisted of six links with 1-DOF \replacedRtwo{pitch-rotational}{revolute} joints.
    Link length and mass match those of the PCS segments.
    Fig.~\ref{fig:walk03_rigid_1} shows the RMSE of the GRF \deletedRtwo{across three phases of gait in the $y$-axis (lateral–medial direction)} between estimates and force-plate measurements.
    \addedRtwo{We selected the GRF in the $y$-axis (lateral–medial direction), $F_y$, because the simplification by the rigid-link model limits its motion in the sagittal plane ($x$-$z$ plane) and therefore mainly results in $F_y$.}
    It is observed that the RMSE of the rigid-link model is higher than that of the hybrid-link model.
    Therefore, this result shows that the hybrid-link system incorporating the PCS model can estimate system accelerations, velocities and forces more accurately than the \replacedRtwo{simplified model}{rigid multi-link system}.
    \addedRtwo{Note that in our analyses the pelvis link is constrained in the sagittal plane; however, the motion dynamics are calculated by 3D model. Therefore, $F_y$ is also a dominant factor in the motion.}
}

\subsection{\addedRone{\replacedRtwo{Elaboration}{Effect of Additional Joints} in Skeletal Model}}
\addedRone{
We\replacedRtwo{
also evaluate the effect of elaborating}
}
{
investigated effect of additional joints in} the skeletal model.
We \replacedRtwo{prepared 11 DOFs skeletal model with additional}{added the} left and right hip \addedRtwo{roll (}abduction–adduction\addedRtwo{)} joints,
\deletedRtwo{in the skeletal model (total 11 DOFs)} trained the policy by RL and 
\replacedRtwo{compared the GRF estimation with that by the original 9 DOFs skeletal model}{
simulate the walking motion}.
Fig.~\ref{fig:grf_walk} shows the \replacedRtwo{RMSE of $F_y$ in a manner similar to Fig.~\ref{fig:walk03_rigid_1} because the additional roll joints should mainly affect $F_y$}{vertical GRF in the simulations, in which two lines indicate the results using the 11 DOFs model and the original 9 DOFs model, respectively}.
\deletedRtwo{
The RMSE is 11.8\%, which implies that adding joints did not have significant effect on the motion analysis setting used in this study.}
\addedRtwo{As shown in the figure, the RMSE using the 11 DOFs model is smaller than that using the 9 DOFs model in the landing phase; however, the RMSE increases in the push-off phase.
In total, the RMSE obtained using the 11 DOFs model is comparable to that obtained using the 9 DOFs model, suggesting that increasing the complexity of the skeletal model does not significantly affect the accuracy of the GRF estimation.
In summary, simplification of the prosthesis model degrades GRF estimation accuracy, whereas increasing the complexity of the skeletal model has little effect. However, increasing the complexity of the skeletal model remains important for enabling three-dimensional analysis in future work.
}


\subsection{\addedRone{Sensitivity Analysis on Contact Model Parameter}}
\addedRone{
We analyzed the sensitivity of the ground contact model parameter for GRF and stance timing.
Fig.~\ref{fig:contact_time} shows the GRF and stance timing when changing the spring coefficient \addedRtwo{of the spring-damper model} by $\pm 10\%$ and $\pm 20\%$ from its nominal value ($8.0$ kN/m).
The mean and standard deviation of stance time are 0.214 and 0.01, respectively.
These results indicate that such variations in the contact model parameter do not significantly affect whole-body motion.}

\section*{ACKNOWLEDGMENT}
The authors sincerely thank Sunghee Kim and Takuma Akiyoshi for contributions to discussions in the Appendix.





\bibliographystyle{IEEEtran}
\bibliography{IEEEabrv,ref}

\end{document}